\documentclass{article} % For LaTeX2e
\usepackage{iclr2024_conference,times}

\usepackage{amsmath,amsfonts,bm}

\def\eqref#1{equation~\ref{#1}}
\def\1{\bm{1}}

\DeclareMathAlphabet{\mathsfit}{\encodingdefault}{\sfdefault}{m}{sl}
\SetMathAlphabet{\mathsfit}{bold}{\encodingdefault}{\sfdefault}{bx}{n}

\usepackage{hyperref}
\usepackage{amsmath}
\usepackage{url}
\usepackage{listings, booktabs}
\usepackage{graphicx}
\usepackage{subcaption}
\usepackage{cleveref}

\usepackage{wrapfig}
\usepackage{algorithm}
\let\algorithmicrequire\relax
\let\algorithmicensure\relax

\usepackage{algpseudocode}

\usepackage{multicol, multirow}
\usepackage{xcolor}
\usepackage{tabularx}
\usepackage{booktabs}
\usepackage[labelfont=bf,format=plain,justification=raggedright]{caption}
\usepackage{arydshln}

\usepackage{amsthm}

\theoremstyle{definition}
\newtheorem{definition}{Definition}[section]

\definecolor{codegreen}{rgb}{0,0.6,0}
\definecolor{codegray}{rgb}{0.5,0.5,0.5}
\definecolor{codepurple}{rgb}{0.58,0,0.82}
\definecolor{backcolour}{rgb}{0.95,0.95,0.92}

\lstdefinestyle{mystyle}{
  backgroundcolor=\color{backcolour}, commentstyle=\color{codegreen},
  keywordstyle=\color{magenta},
  numberstyle=\tiny\color{codegray},
  stringstyle=\color{codepurple},
  basicstyle=\ttfamily\footnotesize,
  breakatwhitespace=false,         
  breaklines=true,                 
  captionpos=b,                    
  keepspaces=true,                 
  numbers=left,                    
  numbersep=5pt,                  
  showspaces=false,                
  showstringspaces=false,
  showtabs=false,                  
  tabsize=2
}

\usepackage[strict]{changepage}
\usepackage{xcolor}
\usepackage{framed}
\definecolor{demonstrationshade}{rgb}{0.95,0.95,1}
\definecolor{promptshade}{rgb}{0.95,0.95,1}

\title{Semantic Scheduling for LLM Inference}

\author{
Wenyue Hua\thanks{Wenyue Hua and Dujian Ding contribute equally. Correspondence to wenyuehua@ucsb.edu} \\
University of California, Santa Barbara \\
\And
Dujian Ding\footnotemark[1] \\
University of British Columbia \\
\And
Yile Gu \\
University of Washington, Seattle \\
\And
Yujie Ren \\
EPFL \\
\And
Kai Mei \\
Rutgers University, New Brunswick \\
\And
Minghua Ma \\
Microsoft \\
\And
William Yang Wang \\
University of California, Santa Barbara \\
}

\begin{document}

\maketitle

\begin{abstract}
Conventional operating system scheduling algorithms are largely content-ignorant, making decisions based on factors such as latency or fairness without considering the actual intents or semantics of processes. Consequently, these algorithms often do not prioritize tasks that require urgent attention or carry higher importance, such as in emergency management scenarios. However, recent advances in language models enable semantic analysis of processes, allowing for more intelligent and context-aware scheduling decisions. In this paper, we introduce the concept of \emph{semantic scheduling} in scheduling of requests from large language models (LLM), where the semantics of the process guide the scheduling priorities. We present a novel scheduling algorithm with optimal time complexity, designed to minimize the overall waiting time in LLM-based prompt scheduling. To illustrate its effectiveness, we present a medical emergency management application, underscoring the potential benefits of semantic scheduling for critical, time-sensitive tasks\footnote{The code and data are available at \url{https://github.com/Wenyueh/latency_optimization_with_priority_constraints}}.
\end{abstract}

\section{Introduction}
Large language models (LLMs) are increasingly prevalent in a variety of domains, serving millions of users worldwide \citep{yu2024aipatient, atkinson2020explanation}. Recent efforts to enhance LLM performance have focused on efficient serving architectures \citep{kwon2023efficient, dao2022flashattention, hua2024interactive}, with the primary objectives of lowering latency and enhancing throughput. However, as LLM applications expand into areas such as medicine \citep{yu2024aipatient} and law \citep{atkinson2020explanation}, it becomes clear that the semantics \citep{mei2024aios} of each request (\emph{e.g.}, the urgency or importance of the request content) can be critical to scheduling decisions.

Most LLM services currently employ a first-come-first-served (FCFS) scheduling strategy, largely because the running time for each user request is unknown. This approach, however, frequently results in head-of-line blocking, as each subsequent request must wait for the current request to complete. Recent methods \citep{fu2024efficient, jin2023s} have addressed this challenge by using smaller language models to predict the expected generation length bucket or generation length ranking of user requests, thereby enabling scheduling strategies akin to shortest-job-first. While such techniques can improve latency and throughput, they do not account for the increasing diversity of LLM applications, where request semantics can significantly influence scheduling priorities.

For example, in emergency and critical response systems that leverage LLMs --- such as healthcare applications \citep{fan2024EMS, yu2024aipatient, gebreab2024llm}, defense \citep{mikhailov2023optimizing, caballero2024large}, or disaster management \citep{otal2024llm, gupta2024utilizing} --- responding promptly to high-urgency requests is paramount. Delays in processing critical requests can have severe consequences, including loss of life or property damage. In these contexts, semantic scheduling, which inspects the content of user requests to assign priority levels, is essential \citep{lin2024parrot}. This represents a departure from conventional scheduling algorithms which remain fundamentally \emph{content-ignorant}: they have no knowledge of the content and logic of each job and consequently rely on metrics such as CPU timeslices \citep{xu2012vslicer, kim2020reconciling} or basic user-defined priorities \citep{saleem2000simulation}.

Semantic attributes of LLM user requests may include urgency \citep{zhu2023priority}, dependencies \citep{grandl2016graphene}, and more. In this paper, we focus on the dimension of urgency within an emergency management environment. We aim to optimize overall latency (\emph{i.e.} the waiting time between the time interval between the arrival time and the completion time of a request) while adhering to the following rule: any highly urgent request should be completed as quickly as possible, even if it requires preempting a currently running request of lower urgency. Towards this end, we propose a new scheduling algorithm that processes requests and evict KV cache memory by priority then by estimated remaining running time, thereby minimizing waiting time for high-priority requests and ensuring that system resources are allocated to those requests where timing is most critical. By integrating semantic information, particularly urgency, into LLM scheduling policies, we provide a framework that is more responsive to real-world needs in high-stake settings. 

Our system reduces latency by leveraging multiple operations to optimize scheduling and resource allocation:\\

% \textbf{Preemption} The system interrupts the ongoing request if a higher-urgency request arrives or if a newly arrived request requires less processing time \citep{zhao1987preemptive}. This ensures that more critical or shorter tasks receive immediate attention and the overall waiting time is minimized.\\

\textbf{Priority-aware scheduling} ensures that requests with high semantic importance bypass lower-priority tasks, reducing head-of-line blocking. \\
\textbf{Stage-aware continuous batching} prevents priority inversions between prefilling and decoding stages \citep{zhong2024distserve}. By dynamically adjusting batch composition based on the highest-priority request's computational stage, the system ensures that urgent decoding requests are not delayed by lower-priority prefilling operations.\\
\textbf{Adaptive memory management} through priority-based eviction maximizes the effective use of limited GPU memory. The dual-heap architecture enables rapid identification of eviction candidates—requests with low semantic importance and high remaining computation time.\\
\textbf{Trade-Off Between Cache Saving and Recomputation} Our work extends traditional OS job scheduling by Key-Value (KV) cache management \citep{lee2024infinigen} in prefilling and decoding steps unde memory eviction: the system determines the optimal amount of previously computed KV (including prefill and decoded tokens) to retain or discard upon eviction. By selectively preserving computation, the system minimizes additional processing time when switching between requests.\\
\textbf{Asynchronous Queue Management} The system asynchronously sorts both queued and newly arriving requests, dynamically determining the next request to process. The scheduling order is updated in real-time as new requests arrive, ensuring that high-priority tasks are consistently prioritized for execution.

By integrating these mechanisms, our system continuously adapts to incoming requests and dynamic priority shifts, minimizing latency, maximizing computational resource efficiency, and ensuring that the most critical requests are prioritized for timely processing.

\section{Related Work}

\paragraph{LLM serving system} A variety of LLM serving systems have been developed to enhance inference speed and efficiency. Many different approaches have been proposed for faster inference \citep{wan2023efficient,ding2024hybrid,ding2025occam} such as algorithm-level inference acceleration including speculative decoding \citep{leviathan2023fast, liu2023online}, KV cache optimization \citep{shi2024keep, liu2024minicache} and system-level inference acceleration including FlashAttention \citep{dao2022flashattention} and PagedAttention \citep{kwon2023efficient}. Many frameworks are developed such as DeepSpeed \citep{rasley2020deepspeed}, Megatron \citep{shoeybi2019megatron}, Colossal-AI \citep{li2023colossal}, vLLM, TensorRT-LLM, etc.

\paragraph{LLM scheduling} Though scheduling of LLM request is a relatively under-explored field, several works worth noticing. \citep{wu2023fast} uses preemptive scheduling to minimize latency with a novel skip-join Multi-Level Feedback Queue scheduler. Based on the semi-information-agnostic setting of LLM inference, the scheduler leverages the input length information to assign an appropriate initial queue for each arrival job to join. To solve the issue of unknown processing time due to unknow generation length\citep{fu2024efficient} proposes to utilize a small language model to predict generation length rankings to achieve lower latency using Shortest Job First scheduling; \citep{kwon2023efficient} instead proposes to predict the generation length bucket. \citep{liu2024andes} introduces a novel quality of experience (QoE) metric for online text services, which measures human satisfaction during the whole token delivery. \citep{Shahout2024fast} proposes a scheduling that both minimize memory waste as well as starvation prevention techniques and optimizations to mitigate the overhead of our scheduling. 

In this paper, we propose \textit{semantic scheduling}, where language models analyze key semantic attributes of user requests, such as urgency and importance, based on \textit{domain-specific} requirements. This content-aware approach goes beyond latency-based scheduling, ensuring high-priority requests receive timely processing in LLM-based services.
\vspace{-10pt}

\section{Semantic Scheduling}
\label{sec:semantic_scheduling}
Semantic scheduling refers to scheduling decisions that are informed by the content or meaning of the task or process at hand. This approach is especially critical in high-stakes scenarios where prioritizing certain tasks can mean preventing severe consequences or saving lives.

\subsection{Motivation by Example}
We illustrate the necessity of semantic scheduling by examining two representative works, \citep{fan2024EMS} and \citep{otal2024llm}, which employ LLMs in high-stakes applications. Both studies provide compelling evidence that traditional scheduling approaches that focus solely on time or resource efficiency are insufficient when the semantics of user requests are critical to system outcomes.

\paragraph{Emergency Medical Services}
~\citep{fan2024EMS} explores the use of LLMs in emergency medical services (EMS). Here, LLMs function as call distributors for emergency department center, parsing incoming patient or caller information to determine the severity of a given situation and routing calls to the appropriate department. Since some requests involve life-threatening conditions while others are less critical, effective scheduling must account for both the timing and the semantic content (\emph{e.g}., symptoms, location details, vital signs) of each call. 

A typical Emergency Severity Index (ESI) \citep{sax2023evaluation} ranks patient conditions as follows: Level 1 - Immediate: Life-threatening; Level 2 – Emergency: Potentially life-threatening; Level 3 – Urgent: Not life-threatening, but requires prompt attention; Level 4 – Semi-urgent: Less critical, intervention can be delayed; Level 5 – Non-urgent: Can be handled as time permits. Given the diversity in urgency, distributing computational resources to ensure that Level 1 or 2 calls receive immediate processing is paramount. By embedding semantic analysis within the scheduling mechanism, the system can reliably differentiate between a patient in acute distress and a routine inquiry, thereby reducing potential delays in high-risk cases.

\paragraph{911 Dispatcher Management}
~\citep{otal2024llm} examines crisis management of 911 dispatch with LLMs focusing on efficiency. Their system processes large, continuous streams of social media posts and real-time messages, with the goal of detecting threats, natural disasters, service disruptions, and infrastructural or social vulnerabilities. Upon receiving a message, a small AI model performs real-time transcription of spoken emergency calls, converting them into text and subsequently segmenting this text for named entity recognition, which extracts key information such as the type and level of emergency, contact details, and other relevant data, thereby functioning as a semantic parser for 911 dispatcher requests. Based on the transcribed and parsed information, the LLM then generates assistive instructions for emergency responders or dispatchers, relaying essential data to the appropriate response systems. Given the wide variation in severity across incoming messages, an effective scheduling algorithm is necessary to rapidly identify high-priority incidents and allocate sufficient computational resources to address them \footnote{Unfortunately, we are not able to experiment on this dataset which is not publicly available}.

\vspace{-10pt}
\paragraph{Advantages of Semantic Scheduling}
In both examples of EMS and broader emergency management, semantically informed scheduling becomes essential. Traditional first-come-first-served (FCFS) or shortest-job-first (SJF) strategies may inadvertently delay life-critical requests if those requests happen to arrive alongside a surge of less urgent tasks. By contrast, semantic scheduling: (1) Identifies High-Stakes Content: Pinpoints urgent messages and prioritizes them based on domain-specific criteria (\textit{\emph{e.g}.}, ESI levels). (2) Allocates Computation Resource: Dynamically decides which tasks to process first when faced with limited GPU or CPU availability. (3) Mitigates Consequences of Delays: Ensures that messages indicating severe threats or emergencies receive the highest level of attention, reducing the potential for catastrophic outcomes.

\subsection{Optimization Setup}
In this work, we interpret the semantics as different \emph{levels of emergency}, which can be framed as priority-aware scheduling \citep{zhu2023priority, hussin2011priority, kundan2021priority}. For each incoming user request, a smaller language model first assigns an emergency level according to some predefined standard, and then the primary LLM processes the request fully. By using these assigned emergency levels, the set of user requests $P$ naturally forms a \emph{graded partially ordered set (poset)}:

\begin{definition}[Graded Poset on User Requests]
A poset ($P$, $\leq$) is called a graded poset if there exists a ranking function $\rho: P\to \mathbb{N}$ where $\rho$ satisfies the following properties: (1) $\rho(p_i) = 0$ iff $\not\exists p_j\in P$ such that $p_j < p_i$ (2) $\forall p_i, p_j\in P$, $\rho(p_j) = \rho(p_i) + 1$ iff $p_i < p_j$ and $\not\exists p_l\in P$ such that $p_j < p_l < p_i$ (3) $\rho$ is compatible with the ordering: $p_i < p_j$ iff $\rho(p_i) < \rho(p_j)$
\end{definition}

%Graded poset has 2 key properties that we need to consider: (1) Consistent Maximal Chain Lengths: For every pair of elements $x\leq y$ in the poset, all maximal chains from $x$ to $y$ must have the same length. (2) The poset can be visualized in layers or tiers, where each layer consists of elements of the same rank. This layering reflects the hierarchical structure imposed by the rank function.

As a concrete example, in the aforementioned EMS application, the ranking function $\rho$ can be defined with respect to the classified emergency level.
Specifically, an LLM call $p_i$ classified as ``level 1 - Immediate: Life-threatening'' has $\rho(p_i) = 0$, while other LLM calls $p_j$ of level $k$ have $\rho(p_j) = k-1$. Clearly, the LLM calls of the same urgency levels form a same layer and all maximal chains between two given LLMs calls have the same length.

\begin{figure*}
    \centering
    \includegraphics[scale=0.5]{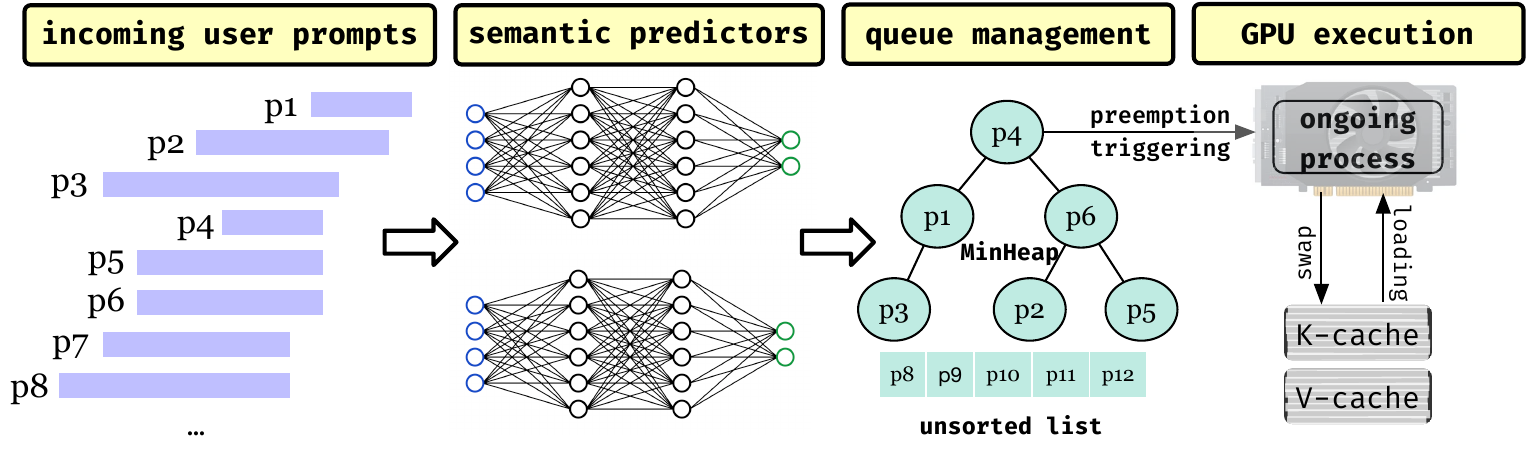}
    \vspace{-10pt}
    \caption{Semantic-Aware Scheduling Pipeline for LLM Inference. Incoming user prompts are processed by a semantic predictor (for urgency) and an output length predictor (for computational cost). Requests are then managed in a MinHeap, with new arrivals stored in an unsorted list before insertion. Preemption triggering determines if a new request should interrupt the ongoing GPU execution, which dynamically handles KV cache saving/loading/swapping to optimize performance and reduce latency.}
    \vspace{-10pt}
    \label{fig:main}
\end{figure*}

\paragraph{Optimization Goal.} We now define our scheduling objective under the assumption that user requests $P$ arrive sequentially over time. The system allows \emph{preemptive} interruption of ongoing processes, ensuring that a request $p_i$ is finished before another request $p_j$ if and only if $p_i$ has a higher ordering (\emph{i.e.}, higher priority or rank) than $p_j$, or $p_i$ completes before $p_j$even arrives.

Formally, let $a_i$ denotes the arrival time of $p_i$ and $f_i$ denotes the finishing time of $p_i$. We aim to minimize the \emph{average waiting time}:
\begin{equation}
    \frac{1}{T}\sum\limits_{i=0}^{i=T} f_i - a_i
\end{equation}
\vspace{-10pt}

subject to the \emph{relative constraint}:
\vspace{-10pt}
\begin{multline}
    \forall p_i\in P, \forall p_j\in P, f_i < f_j\to\\
    \ (f_i < a_j)\lor (r(p_i) <= r(p_j))
\end{multline}
which encodes that if a request $p_i$ is completed before another request $p_j$, then $p_i$ was completed before the arrival of $p_j$', or $p_i$ has a higher (or equal) priority according to the ranking function $r$.
This formulation reflects our goal of respecting emergency-based semantics: urgent requests must be prioritized even if preempting ongoing work is necessary. By enforcing these constraints and optimizing overall waiting time, we seek to deliver prompt service to the most critical user requests while maintaining efficient resource usage.
\vspace{-5pt}

\subsection{Algorithm}
% \begin{algorithm}[!ht]
% \DontPrintSemicolon
% \KwIn{A batch of requests $P$, number of top requests $k$}
% \KwOut{Indices of $k$ highest-ranked requests}
% $\textsc{TopKList} \gets$ empty list of size at most $k$

% \For{$i \gets 0$ \KwTo $|P|-1$}{
%     $u \gets f_{urgency}(P[i])$\;
    
%     $t \gets f_{time}(P[i])$\;
    
%     $\textsc{priority} \gets (u, t)$\;
    
%     \If{$\textsc{TopKList.size} < k$}{
    
%         \textsc{TopKList.push}($P[i]$)\;
        
%     }
% }
% \Return \textsc{TopKList}
% \caption{FindTopK \ddj{TopKList comprises the first k requests in P. Seems not essentially the top-k.}}
% \label{algo:findtopk}
% \end{algorithm}

In the proposed approach, each arriving user request (or batch of requests) undergoes two preliminary prediction steps: (1) Emergency-Level Prediction: A small language model determines the priority or urgency of the request. (2) Output Generation Length Prediction: A length-bucket predictor using the $S^3$ model from \citep{jin2023s} provides an approximate estimate of the request's total processing time, factoring in both prefill time for the input prompt and decoding time for the predicted output length.

\begin{algorithm}
\caption{MinHeap Construction}
\begin{algorithmic}[1]
\Require MinHeap $\mathcal{H}$, unsorted buffer $\mathcal{U}$, incoming requests $\mathcal{R}$
\Ensure Updated MinHeap $\mathcal{H}$ and buffer $\mathcal{U}$

\While{system is active}
    \State $\mathcal{U} \gets \mathcal{U} \cup \mathcal{R}$ \Comment{Append to unsorted buffer}
    \If{$|\mathcal{U}| > 0$}
        \For{each request $r \in \mathcal{U}$}
            \State $r.priority \gets$ \textsc{ObtainPriority}$(r.f_e, r.f_t)$
            \State \textsc{HeapInsert}$(r, \mathcal{H})$ \Comment{$\mathcal{O}(\log n)$}
            \State $\mathcal{U}$.remove($r$)
        \EndFor
        
    \Else
        \State Sleep$(\delta t)$ \Comment{Wait until new requests coming if necessary}
    \EndIf
\EndWhile

\State \textbf{return} $\mathcal{H}, \mathcal{U}$

\end{algorithmic}
\end{algorithm}

With these predictions, the system makes scheduling decisions to minimize overall waiting time while respecting priority constraints. The following sections detail the algorithm's key components: build an ordered heap of requests based on incoming requests asynchronoulsy, scheduling the next batch to run for each iteration, handling memory caching and recomputation. The main process is described in Figure \ref{fig:main}.

\paragraph{Asynchronous Queue Management}

In resource-constrained environments, incoming requests may accumulate faster than the system can process them, particularly under high-load conditions. To address this challenge, we implement an asynchronous queue management system using a minimum-heap data structure. Each request in the queue is prioritized according to multiple scheduling criteria: an emergency level parameter $f_e$ and the estimated remaining computation time $f_t$.

Our approach decouples queue management from GPU execution through asynchronous processing. While the GPU processes the current batch of requests, a separate asynchronous process maintains the priority queue by incrementally inserting newly arrived requests into the heap. This design eliminates blocking between request arrival and GPU processing, enabling continuous queue updates without interrupting ongoing computations.

The system maintains two distinct data structures: a minimum heap for prioritized requests ready for scheduling, and an unsorted buffer for newly arrived requests awaiting insertion. The asynchronous process continuously transfers requests from the buffer to the heap, with each insertion operation requiring $\mathcal{O}(\log n)$ time, where $n$ denotes the current heap size. This incremental approach avoids the computational overhead of periodic full-heap reconstructions while ensuring that the highest-priority request can be efficiently retrieved in $\mathcal{O}(1)$ time when the GPU becomes available for the next batch.

%When the current request is done and we need to choose the next request to run but the heap remains partially constructed, the system compares the highest-priority request in the heap with the top-priority request from the unsorted list by \textsc{FindMax}, which incurs a computational cost of approximately $\mathcal{O}(n') + \mathcal{O}(1)$ where $n'$ is the number of pending requests in the unsorted list. 

\paragraph{Schedule the Highest-Priority Batch Requests}
At each iteration of computation, whether a prefilling iteration or a single-token decoding iteration, the scheduler selects the $b$ highest-priority requests for execution. This selection process examines both the built MinHeap and any requests remaining in the unsorted buffer that have not yet been inserted into the heap, ensuring that no high-priority request is overlooked due to asynchronous queue management delays. The scheduler extracts the top-$b$ requests based on their scheduling priority tuple $(f_e(p), f_t(p))$.

The selected requests must then be merged with ongoing requests from the previous iteration. This merging process requires careful handling due to the continuous batching paradigm, where prefilling and decoding operations are processed separately within each batch. Specifically, when a batch contains requests in different stages, the system prioritizes prefilling operations: all requests requiring prefilling are processed first, followed by requests in the decoding stage. This stage-based processing can create a priority inversion problem in which lower priority prefilling requests block higher priority decoding requests from execution.

\begin{algorithm}
\caption{Scheduling with Stage Awareness}
\label{alg:batch_scheduling}
\begin{algorithmic}[1]
\Require MinHeap $\mathcal{H}$, unsorted buffer $\mathcal{U}$, ongoing requests $\mathcal{O}$, batch size $b$
\Ensure Next batch of requests to process

\State $C \gets \textsc{ExtractTopb}(\mathcal{H}, \mathcal{U}, b)$ \Comment{Extract $b$ highest-priority candidates}

\State $p^* \gets \arg\max_{p \in C \cup \mathcal{O}} \textsc{Priority}(p)$ \Comment{Find highest-priority request $p^*$}

\If{$\textsc{Stage}(p^*) = \texttt{PREFILLING}$}
    \State $M \gets \textsc{Merge}(C, \mathcal{O})$ \Comment{Merge all requests}
\Else 
    \State $C_{decode} \gets \{p \in C : \textsc{Stage}(p) = \texttt{DECODING}\}$
    \State $C_{prefill} \gets C \setminus C_{decode}$ \Comment{Separate prefilling requests}
    \State $M \gets \textsc{Merge}(C_{decode}, \mathcal{O})$ \Comment{Merge only decoding requests}
    \State \textsc{PushBack}$(C_{prefill}, \mathcal{H})$ \Comment{Push back prefilling requests to $\mathcal{H}$}
\EndIf

\State $B_{next} \gets M[:b]$ \Comment{Select top-$b$ requests}
\State $M_{unselected} \gets M[b:]$ \Comment{Remaining unselected requests}
\State \textsc{PushBack}$(M_{unselected}, \mathcal{H})$ \Comment{Push back unselected requests back to $\mathcal{H}$}

\State \textbf{return} $B_{next}$

\end{algorithmic}
\end{algorithm}

To address this issue, we implement a stage-aware scheduling algorithm. After selecting the top-$b$ requests, the scheduler identifies the highest-priority request $p^*$ among them. If $p^*$ is in the prefilling stage, the scheduler proceeds with standard merging and batch formation. However, if $p^*$ is in the decoding stage, the scheduler filters out all newly selected prefilling requests, merging only those requests that are also in the decoding stage. This approach ensures that high-priority decoding requests are not delayed by lower-priority prefilling operations, maintaining the integrity of the priority-based scheduling while respecting the computational constraints of continuous batching.

After selecting the top-$b$ requests, other unselected requests will be pushed back to the MinHeap for future scheduling iterations.This ensures that no request is lost during the selection process and maintains the priority ordering of all pending requests in the system. The pushback operation preserves the heap property, allowing these unselected requests to be reconsidered in subsequent scheduling rounds based on their priority values $(f_e, f_t)$.

\paragraph{KV Cache Management}
During request computation, we adopt commonly-used key-value (KV) cache for each processed request in GPU memory to enable efficient generation. When on-device memory approaches capacity, we implement a priority-based eviction strategy that removes KV data of requests with the lowest scheduling priority determined by their semantic importance $f_e$ and estimated remaining computation time $f_t$, which we name as victim requests. This approach ensures that high-priority requests maintain low latency by preserving their cached states by sacrificing requests with low scheduling priority.

\begin{algorithm}
\caption{MaxHeap Construction}
\begin{algorithmic}[1]
\Require Array of active requests $\mathcal{R}$ with KV caches, MaxHeap $\mathcal{G}$ ordered by eviction priority
\Ensure MaxHeap $\mathcal{G}$ ordered by eviction priority

\For{each request $r \in \mathcal{R}$}

    \State $r.eviction\_priority \gets$ \textsc{ObtainPriority}$(- r.f_e, - r.f_t)$ \Comment{Compute eviction priority for each request}
    
    \State \textsc{HeapInsert}$(r, \mathcal{G})$ 
\EndFor

\State \textbf{return} $\mathcal{G}$

\end{algorithmic}
\end{algorithm}

To enable efficient cache eviction, we maintain a maximum-heap data structure that contains all active KV cache entries currently resident on the device. This eviction heap ranks requests with cache in GPU inversely to the scheduling heap: requests with the lowest semantic priority $f_e$ and longest remaining computation time $f_t$ appear at the heap's root, making them immediate candidates for eviction. 

This design creates a complementary relationship between our two heap structures: the scheduling min-heap prioritizes requests with high $f_e$ and low $f_t$ for execution, and the eviction max-heap identifies requests with low $f_e$ and high $f_t$ for memory reclamation. The dual-heap architecture provides fine-grained control over both computational scheduling and memory management. When memory pressure occurs, we can efficiently identify and evict the least critical KV caches in $\mathcal{O}(\log n)$ time, where $n$ represents the number of cached requests. This mechanism ensures optimal system responsiveness under memory-constrained conditions by maintaining cache residency for requests most likely to complete soon or those with highest semantic importance, while gracefully degrading service for lower-priority requests through selective eviction.

\paragraph{Cache and Recomputation Strategy}
Once some cache is evicted, the system evaluates whether it is more efficient to discard the cache entirely or to recompute the corresponding part when the request resumes based on our predefined Cache and Recomputation Strategy. Then, the affected request's position in the scheduling queue is also updated to reflect its revised remaining computation time.

When a request's KV is evicted, the scheduler must decide how many computed tokens' KV cache should be saved and how many to discard for later recomputation in order to minimize latency. This decision depends primarily on (1) how far the request has progressed (\emph{i.e.}, whether it is more efficient to retain already computed data) and (2) whether recomputation is cheaper than incurring the overhead of cache saving and loading. To clarify this trade-off, consider the approximate processing and cache-loading times for the prefill and decoding stages:

For a given prompt of length $n$, we estimate the prefilling time using the quadratic model $\alpha_1 n^2 + \alpha_2 n$, where the quadratic part accounts for the time spent computing attention mechanism and the linear part accounts for the non-negligible time spent on the feedforward network. The constants $\alpha_1$ and $\alpha_2$ are specific to the employed model and serving engine. If this request is evicted and the entire prefilling KV cache is offloaded to CPU, reloading it approximately takes $\beta n$, where $\beta$ is the cache loading speed. For now, we do not consider advanced KV cache optimization techniques \citep{shi2024keep} and consider that the loading time scales linearly with the length of prompt.

If the evicted cache include $n$ tokens' KV cache done in the prefilling time, the system must decide whether to offload them to the CPU or to simply recompute the prefill upon resumption of the request later. To minimize latency, it is necessary to determine a threshold for deciding between caching and recomputation. Specifically, if
\begin{equation}
    \beta n > \alpha_1 n^2 + \alpha_2 n, \text{ \emph{i.e.} } \beta >  \alpha_1 n + \alpha_2
\end{equation}
\vspace{-20pt}

, then caching is more efficient; otherwise, recomputation is preferable. This threshold also applies when the prompt are only partially prefilled.

For a given prompt of length $n$ for which the system aims to decode $m$ additional tokens, to decode the $m'$th token given the computed prompt and $m'-1$ decoded tokens, the decoding time can be approximated by $\gamma_1 (n + m'-1) + \gamma_2$, where $\gamma_1$ captures the linear cost of attention (assuming reuse of KV caches) and $\gamma_2$ captures the constant feedforward computation needed for each decoded token. Consequently, the total cost of decoding $m$ tokens is approximated by summing over the tokens from $n+1$ to $n+m$: 
\begin{equation}
    \sum_{i = n+1}^{i = n+m}(\gamma_1 i + \gamma_2) = \gamma_1(\frac{1}{2} m^2 + n m + \frac{1}{2} m) + \gamma_2 m
\end{equation}
\vspace{-10pt}

\begin{algorithm}
\caption{Priority-Based Eviction}
\begin{algorithmic}[1]
\Require Request $r$, MaxHeap $\mathcal{G}$ for current KV cache, MinHeap $\mathcal{H}$ for waiting queue, DeviceCacheCapacity $C$
\Ensure MaxHeap $\mathcal{G}$, MinHeap $\mathcal{H}$

\State RequiredKV $s$ $\gets$ \textsc{EstimateKVSize}($r$)

\State CurrentKVUsage $c$ $\gets$ \textsc{ComputeCurrentKVCache}($\mathcal{G}$)

\While{$s + c > C$}
    \State VictimRequest $r_v$ $\gets$ $\mathcal{G}$.pop() \Comment{choose request to evict based on priority}
    \State $\mathcal{H}$.delete($r_v$)
    \Comment{remove $r_v$ from $\mathcal{H}$}
    \If{$\textsc{ShouldRecompute}(r_v)$} 
    
        \State \textsc{DiscardKVCache}$(r_v)$ 
        \Comment{if recomputing $r_v$ is faster, then discard}

    \Else
        
        \State \textsc{OffloadToCPU}$(r_v)$
        \Comment{if recomputing $r_v$ is slower, then offload}
    \EndIf

    \State \textsc{UpdateRemainingComputationTime}$(r_v)$

    \State $\mathcal{H}$.push($r_v$)
    \Comment{update the position of $r_v$ based on updated remaining computation time}

    \State $c$ $\gets$ \textsc{ComputeCurrentKVCache}($\mathcal{G}$)
\EndWhile

\State \textbf{return} $\mathcal{G}, \mathcal{H}$

\end{algorithmic}
\end{algorithm}

If the evicted cache include $m'$ tokens' KV cache done in the decoding time, the system determines the number $m'_*$ of computed tokens' KV cache (where $0\leq m'_* \leq m'\leq m$) to save. Reloading that cache on resumption takes $\beta m'$ units of time. The remaining $m'-m'_*$ tokens will then be recomputed, whose overall decoding cost after preemption is approximately
\begin{multline}
\beta m'_* + \gamma_1(\frac{1}{2} (m'-m'_*
    )^2 + n (m'-m'_*) + \frac{1}{2} (m'-m_*)) + \gamma_2 (m'-m'_*)
\end{multline}

One obtains the optimal number of tokens to save, $m'_{*}$, by setting the derivative of this total time expression with respect to $m'$ to 0 and solving for $m'$: 
\vspace{-5pt}
\begin{equation}
    m'_{*} = \max(0, \frac{\beta - \gamma_1 n - (\frac{\gamma_1 + 2\gamma_2)}{2}}{2\gamma_1})
\end{equation}
\vspace{-15pt}

This formula balances the overhead of loading cached tokens against the cost of recomputing tokens, yielding the minimum overall decoding time if a request is preempted during its decoding phase.

A common approach to improving prefilling efficiency is using a radix tree, a compressed trie for caching prefill KV data \citep{zheng2024sglang, yechunkattention, li2024survey}. A radix tree is built on from previously processed requests' KV cache and for a given new request, by identifying the longest cached prefix in the radix tree, the system reduces computations by processing only the non-cached portion of the prompt. However, in emergency management scenarios, shared prefixes among requests are minimal due to diverse phrasing and content. Our analysis of the EMS dataset \citep{yu2024aipatient} shows an average shared prefix of only 0.10 tokens per request and a maximum of 1.84, making the benefits of a radix tree negligible. To avoid additional overhead and complexity, we omit its use in our current algorithm but acknowledge its potential for future integration.

\begin{algorithm}[!ht]
\begin{algorithmic}[1]
\Require Batch size $b$, GPU memory capacity $M_{cap}$
\Ensure Continuous processing of prioritized requests
\State \textbf{Initialize:} MinHeap $\mathcal{H} \gets \emptyset$, MaxHeap $\mathcal{E} \gets \emptyset$
\State \hspace{2.5em} Buffer $\mathcal{U} \gets \emptyset$, ongoing requests $\mathcal{O} \gets \emptyset$
\State \hspace{2.5em} Memory usage $M_{used} \gets 0$

\State
\State \textbf{// Main Thread: Request Reception}
\While{system active}
    \State $R_{new} \gets$ \textsc{GetIncomingRequests}()
    \State $\mathcal{U} \gets \mathcal{U} \cup R_{new}$ \Comment{Non-blocking append}
\EndWhile

\State
\State \textbf{// Async Thread: Heap Maintenance}
\While{system active}
    \If{$|\mathcal{U}| > 0$}
        \For{each $r \in \mathcal{U}$}
            \State $r.priority \gets (r.f_p, r.f_t)$
            \State \textsc{HeapInsert}$(r, \mathcal{H})$ 
        \EndFor
    \EndIf
\EndWhile

\State \textbf{// GPU Thread: Execution and Memory Management}
\While{system active}
    \State \textbf{// Stage-aware scheduling}
    \State $\mathcal{O}\gets$ \textsc{StageAwareSchedule}$(\mathcal{H}, \mathcal{U}, \mathcal{O}, b)$
    \State \textsc{GPUExecuteBatch}$(\mathcal{O}_{next})$
    
    \State \textbf{// Update eviction heap for active requests}
    \For{each $r \in \mathcal{O}_{next}$}
        \State \textsc{PriorityBasedEviction}$(r, \mathcal{G}, \mathcal{H}, C)$
        \State \textsc{MaxHeapConstruction}$(\{r\}, \mathcal{G})$
    \EndFor
    
\EndWhile

\caption{Priority-Aware Request Processing System}
\label{algo:full}

\end{algorithmic}
\end{algorithm}

\paragraph{Summary}
The proposed system integrates four key components to achieve efficient priority-aware request processing under resource constraints: (1) asynchronous queue management with incremental heap construction, (2) stage-aware continuous batching that respects request priorities, (3) priority-based KV cache eviction using a dual-heap architecture, and (4) adaptive cache-or-recompute strategies. This design ensures that semantically important requests receive preferential treatment in both scheduling and memory allocation, while maintaining system stability under high load conditions. Full algorithm is presented in Alg.~\ref{algo:full}.

The complete system operates through two parallel execution paths, as presented in Alg.~\ref{algo:full}. The main thread continuously receives incoming requests and maintains them in an unsorted buffer, while an asynchronous thread incrementally constructs the scheduling min-heap without blocking request arrivals. Simultaneously, the GPU execution thread performs stage-aware batch scheduling, maintains the eviction max-heap for cached requests, and triggers memory reclamation when necessary. This parallel architecture decouples request arrival from processing, enabling the system to handle burst traffic while preserving priority ordering.

\section{Experiment}
We conduct our experiments in two stages: a \emph{simulated} component and a \emph{real-dataset} component. In the simulated experiments, we evaluate how the following factors affect system performance:\\
(1) \textbf{Semantic Predictor Accuracy.} We quantify how precisely the system identifies each request's urgency level. \\
(2) \textbf{Output-Length Bucket Predictor Accuracy.} We investigate the effects of incorrect or imprecise predictions of request output lengths. \\
(3) \textbf{Computation Speed and Maximum Batch Size for Predictors.} We examine how varying the computational settings and batch sizes impact overall waiting time.\\
(4) \textbf{Scheduling Strategies for Predictors.} We compare an immediate-processing approach (predictors run as requests arrive) versus a delayed-batching strategy (predictors wait until a batch is full). 

Furthermore, we test the system under \emph{request spikes}, where the interval between sequential request arrivals can be as low as 0.1 seconds, and each arrival may contain up to 100 concurrent requests. In non-spike settings, the maximum number of concurrent requests at each timestamp is set to 5.

Both simulation and real-data experiments are conduct under three settings: Qwen1.5-4B and Qwen1.5-7B as the base language models on NVIDIA A100 GPUs, along with Qwen1.5-7B on NVIDIA A5000 GPUs. For semantic prediction and output-length bucket estimation, we use DistilBERT \citep{jin2023s}. To ensure realistic simulations, we profile the models and GPUs under various loads to compute key parameters $\alpha_1, \alpha_2, \beta, \gamma_1, \gamma_2$ (details in Appendix\ref{app:profile}). For real-dataset experiments, the algorithm is implemented in vLLM \citep{kwon2023efficient}. For parameters directly under study, such as the speed, batch size, and accuracy of predictor, we systematically vary their values to observe their impacts on system latency.

We present the results of Qwen1.5-4B on A100 of simulation and real-dataset experiment in the main paper, with additional results for the other three settings in Appendix \ref{app:simulation_result}.

\subsection{Simulated Experiments}
\label{subsec:simulated_experiments}

% \begin{figure}[t]
%     \centering
%     \begin{minipage}{0.65\textwidth}
%         \centering
%         \includegraphics[height=.33in]{Figures/Evaluation/legend_urgency.pdf}
%     \end{minipage}
%     \vfill
%     \begin{subfigure}[t]{0.2\textwidth}
%     \includegraphics[height=1.2in]{Figures/Evaluation/simulation_priority.pdf}
%     \subcaption{Urgency.}
%     \label{fig:simulation_priority}
%     \end{subfigure}%
%     ~
%     \begin{subfigure}[t]{0.2\textwidth}
%     \includegraphics[height=1.2in]{Figures/Evaluation/simulation_length.pdf}
%     \caption{Length.}
%     \label{fig:simulation_length}
%     \end{subfigure}
%     \vfill
%     \begin{minipage}{0.45\textwidth}
%         \centering
%         \includegraphics[height=.33in]{Figures/Evaluation/legend_methods.pdf}
%     \end{minipage}
%     \vfill
%     \begin{subfigure}[t]{0.2\textwidth}
%     \includegraphics[height=1.2in]{Figures/Evaluation/simulation_gap0.1_all_a100_4b.pdf}
%     \caption{Gap=0.1s.}
%     \label{fig:simulation_gap01}
%     \end{subfigure}%
%     ~
%     \begin{subfigure}[t]{0.2\textwidth}
%     \includegraphics[height=1.2in]{Figures/Evaluation/simulation_gap1.0_all_a100_4b.pdf}
%     \caption{Gap=1.0s.}
%     \label{fig:simulation_gap099}
%     \end{subfigure}
%     \caption{Simulation results on the influence of (a) semantic and (b) output length predictor accuracy, and (c-d) response to spikes in user requests with urgency level 0.}
% \label{fig:simulation}
% \end{figure}

% Add to preamble:
% \usepackage{subcaption}

% Add to preamble:
% \usepackage{subcaption}

\begin{figure}[t]
    \centering
    
    % First legend
    \includegraphics[height=.5in]{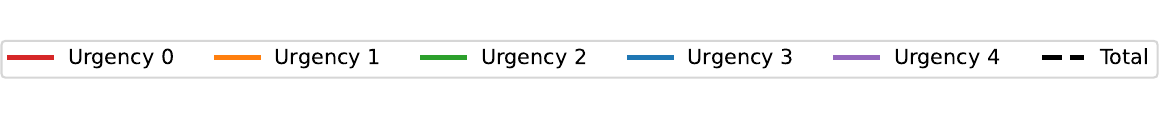}
    
    \vspace{1mm}
    
    % First row of subfigures
    \begin{subfigure}[t]{0.45\textwidth}
        \centering
        \includegraphics[height=2.0in]{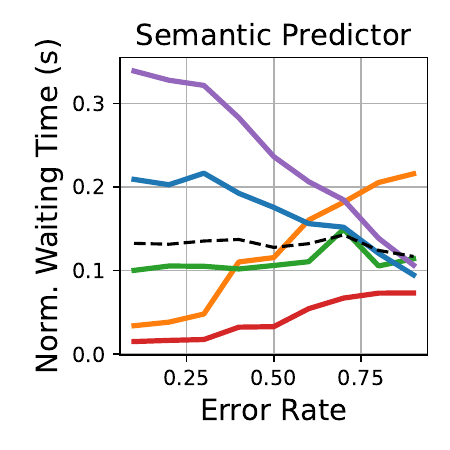}
        \caption{Urgency.}
        \label{fig:simulation_priority}
    \end{subfigure}
    \hspace{5mm}
    \begin{subfigure}[t]{0.45\textwidth}
        \centering
        \includegraphics[height=2.0in]{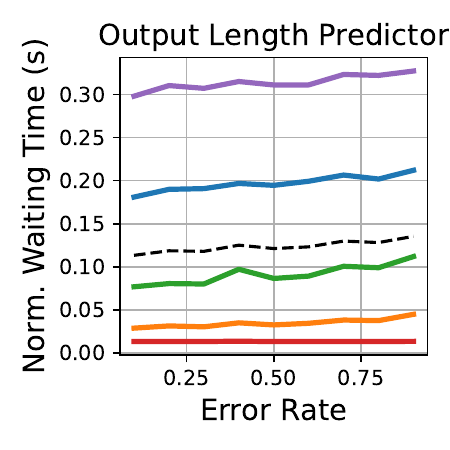}
        \caption{Length.}
        \label{fig:simulation_length}
    \end{subfigure}
    
    \vspace{3mm}
    
    % Second legend
    \includegraphics[height=.5in]{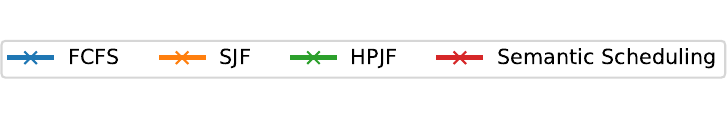}
    
    \vspace{1mm}
    
    % Second row of subfigures
    \begin{subfigure}[t]{0.45\textwidth}
        \centering
        \includegraphics[height=2.0in]{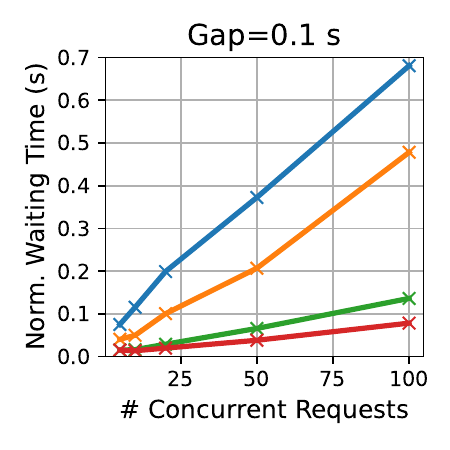}
        \caption{Gap=0.1s.}
        \label{fig:simulation_gap01}
    \end{subfigure}
    \hspace{5mm}
    \begin{subfigure}[t]{0.45\textwidth}
        \centering
        \includegraphics[height=2.0in]{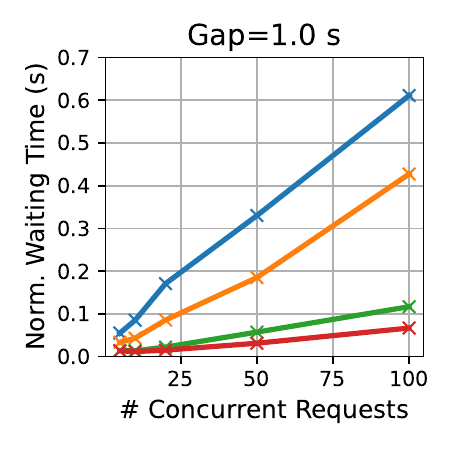}
        \caption{Gap=1.0s.}
        \label{fig:simulation_gap099}
    \end{subfigure}
    
    \caption{Simulation results on the influence of (a) semantic and (b) output length predictor accuracy, and (c-d) response to spikes in user requests with urgency level 0.}
    \label{fig:simulation}
\end{figure}

\paragraph{Influence of Semantic Predictor Accuracy.} 
We investigate the influence of semantic predictor accuracy over the normalized waiting time (the average waiting time per generated token). We use a unified notion of error rate to reflect both the percentage of mistaken requests and the average prediction error (the difference between predicted urgency level and ground truth over the total number of urgency levels). For example, an error rate of 0.2 indicates that the semantic predictor makes wrong prediction for 20\% requests and the average prediction error distance is 0.2. 
We use 5 urgency levels as in ESI and vary error rates from 0.1 to 0.9. A more accurate predictor leads to better discrimination of higher-urgency requests, thus reducing the queue time for critical requests. When accuracy is low, critical requests may be misclassified, resulting in longer queue times for the most urgent tasks and increasing their waiting time. 

Our report results in~\Cref{fig:simulation_priority}. For requests of the highest urgency level (Urgency 0), the normalized waiting time has increased from 0.02s to 0.07s, which is 3.5x slower. Similarly, for requests of urgency level 1, the normalized waiting time increases from 0.03s to 0.22s, indicating a 7.3x slowdown. On contrast, the low urgency requests (Urgency 2,3, and 4) could be mistakenly prioritized that leads to a decreased overall normalized waiting time (the overall normalized waiting time over all urgency levels slightly decreases from 0.13s to 0.12s). It is worth noting that, in practice, high urgency requests may relate to life-threatening dangers (\emph{e.g}., emergency medical service) and require immediate attention. Delays in these high urgency requests are critical and cannot be compensated by the marginal improvements of the overall latency saving.

\paragraph{Impact of Output Length Bucket Predictor Accuracy.} The EMS dataset contains one-round conversations that are relatively short, as in emergency time people usually don't say very long sentences as reply or as prompt. Thus we chunk the length into 5 buckets. and each bucket range around 10 token length. Similar to the semantic predictor, misclassifications in output-length estimation can negatively affect scheduling, as incorrectly anticipated request lengths alter scheduling priorities. We quantify how this affects overall system performance, focusing on both individual request delays and collective wait durations. We adopt a similar notion of error rate to characterize the percentage of mistaken requests and the average prediction error (the difference between predicted output length and ground truth over the maximum output length). We configure the maximum output length to be 500 and vary error rates from 0.1 to 0.9. Results are summarized in~\Cref{fig:simulation_length}. As the error rates increase in the output length predictor, the normalized waiting time increases for all urgency levels.
For the most critical requests (Urgency 0), the normalized waiting time is consistently around 0.01s. For requests of urgency level 1 and 2, the normalized waiting time increases from 0.03s to 0.04 and 0.08s to 0.11s, respectively. For low urgency requests (Urgency 3 and 4), the increase of normalized waiting time is from 0.18s to 0.21s, and from 0.30s to 0.33s separately. The observation demonstrates the importance of developing accurate output length bucket predictor to further accelerate LLM scheduling.

\paragraph{Predictor Computation Speed and Batching.} We examine various predictor configurations, including different speeds and maximum batch sizes. We also compare two types of scheduling in predictor: \emph{Immediate Processing} -- the predictor runs as soon as new requests arrive, and \emph{Full Batching} -- requests are accumulated until reaching a predefined batch size, then processed together. 
Our results reveal trade-offs in terms of overhead and efficiency (see~\Cref{tab:full_batching_immediate_processing_predictor_latency,tab:full_batching_immediate_processing_batch}).
In~\Cref{tab:full_batching_immediate_processing_predictor_latency}, we fix batch size to be 64 and increase predictor latency from 0.001s to 1s. We observe that the normalized waiting time for all urgency levels in general increases for both full batching and immediate processing.
When the predictor latency is low (e.g., 0.001s), immediate processing achieves lower normalized waiting time and outperforms full batching. However, as predictor latency increases, immediate processing becomes less efficient, as it handles and makes predictions for each incoming request individually, whereas full batching processes the entire batch at once, resulting in lower overhead.
In~\Cref{tab:full_batching_immediate_processing_batch}, we consider predictor latency of 0.01 second and study the effect of batch sizes on the normalized waiting time.
For full batching, the normalized waiting time of all urgency levels monotonically increases as the batch size increases.
One possible explanation is that, when request traffic is unsaturated, a large batch size may take long time to fulfill and leads to request response delay.
Immediate processing is observed to be insensitive to batch changes as expected and more efficient than full batching across a majority of experiment settings.
% Overall, while \textit{full batching} may improve throughput, it can also increase queue time for individual requests waiting to be batched and therefore leads to higher normalized waiting time than \textit{immediate processing}.

\begin{table}[]
\centering
\resizebox{0.8\linewidth}{!}{
\begin{tabular}{ccccccccc}
\hline
\begin{tabular}[c]{@{}c@{}}Norm. Waiting \\ Time (s)\end{tabular} &
  \multicolumn{4}{c}{Full Batching} &
  \multicolumn{4}{c}{Immediate Processing} \\ \hline
Predictor Latency &
  {\color[HTML]{333333} 0.001} &
  0.01 &
  0.1 &
  \multicolumn{1}{c|}{1.0} &
  {\color[HTML]{333333} 0.001} &
  0.01 &
  0.1 &
  1.0 \\ \hline
Urgency 0 &
  {\color[HTML]{333333} 0.07} &
  0.09 &
  0.18 &
  \multicolumn{1}{c|}{0.35} &
  0.01 &
  0.08 &
  0.16 &
  0.37 \\
Urgency 1 & 0.13 & 0.35 & 0.68 & \multicolumn{1}{c|}{1.37} & 0.01 & 0.34 & 0.68 & 1.42 \\
Urgency 2 & 0.21 & 0.63 & 1.25 & \multicolumn{1}{c|}{2.51} & 0.02 & 0.62 & 1.24 & 2.55 \\
Urgency 3 & 0.29 & 0.89 & 1.75 & \multicolumn{1}{c|}{3.51} & 0.04 & 0.91 & 1.83 & 3.72 \\
Urgency 4 & 0.37 & 1.15 & 2.29 & \multicolumn{1}{c|}{4.57} & 0.15 & 1.15 & 2.33 & 4.70 \\ \hline
Avg. &
  0.21 &
  \textbf{0.61} &
  \textbf{1.23} &
  \multicolumn{1}{c|}{\textbf{2.46}} &
  \textbf{0.04} &
  \textbf{0.61} &
  1.25 &
  2.56 \\ \hline
\end{tabular}
}
\caption{Normalized waiting time for full batching and immediate processing at different predictor latency levels, when the max number of concurrent requests is 5 and the batch size is 64.}
\label{tab:full_batching_immediate_processing_predictor_latency}
\end{table}

\begin{table}[]
\centering
\resizebox{0.8\linewidth}{!}{
\begin{tabular}{ccccccccc}
\hline
\begin{tabular}[c]{@{}c@{}}Norm. Waiting \\ Time (s)\end{tabular} &
  \multicolumn{4}{c}{Full Batching} &
  \multicolumn{4}{c}{Immediate Processing} \\ \hline
Batch Size & {\color[HTML]{333333} 4}    & 16   & 32   & \multicolumn{1}{c|}{64}   & {\color[HTML]{333333} 4} & 16   & 32   & 64   \\ \hline
Urgency 0  & {\color[HTML]{333333} 0.07} & 0.07 & 0.08 & \multicolumn{1}{c|}{0.08} & 0.06                     & 0.07 & 0.07 & 0.07 \\
Urgency 1  & 0.33                        & 0.34 & 0.35 & \multicolumn{1}{c|}{0.35} & 0.33                     & 0.33 & 0.34 & 0.33 \\
Urgency 2  & 0.60                        & 0.61 & 0.62 & \multicolumn{1}{c|}{0.62} & 0.60                     & 0.60 & 0.60 & 0.60 \\
Urgency 3  & 0.87                        & 0.88 & 0.90 & \multicolumn{1}{c|}{0.91} & 0.87                     & 0.87 & 0.88 & 0.87 \\
Urgency 4  & 1.13                        & 1.14 & 1.17 & \multicolumn{1}{c|}{1.17} & 1.13                     & 1.14 & 1.14 & 1.14 \\ \hline
Total &
  \textbf{0.58} &
  0.59 &
  0.60 &
  \multicolumn{1}{c|}{0.60} &
  \textbf{0.58} &
  \textbf{0.58} &
  \textbf{0.59} &
  \textbf{0.58} \\ \hline
\end{tabular}
}
\caption{Normalized waiting time for full batching and immediate processing with different batch sizes, when the max number of concurrent requests is 5 and the predictor latency is of 0.01s.}
\label{tab:full_batching_immediate_processing_batch}
\end{table}

\paragraph{Response to Spikes in User Requests.} Lastly, we explore scenarios involving high-frequency request arrivals, where gaps can be as short as 0.1 seconds and up to 100 requests may arrive concurrently (see~\Cref{fig:simulation_gap01,fig:simulation_gap099}). We compare our semantic-based scheduling to baselines including First-Come-First-Served (FCFS), Shortest Job First (SJF), and  Highest-Priority Job First (HPJF) for the most critical queries of urgency level 0. Preliminary findings indicate that FCFS, SJF, and HPJF struggle to handle large bursts of concurrent traffic, leading to a sharp increase in normalized waiting time for high-urgency requests, whereas semantic scheduling mitigates these bottlenecks by effectively prioritizing urgent tasks and accurately estimating the request output length.
When request arrival gap is 0.1s, the normalized waiting time of semantic scheduling is up to 0.08s for requests of urgency level 0 with 100 concurrent requests, which are 1.7x, 6.1x, and 8.7x faster compared to HPJF, SJF, and FCFS, respectively. When request arrival gap increases to 1.0s, semantic scheduling achieves up to 9.1x speed-ups with respect to all the baselines. 

Overall, the simulated experiments demonstrate how varying predictor accuracy, computational settings, and batching strategies can significantly affect request prioritization and overall system responsiveness.

\subsection{Real Dataset Experiment}
\label{subsec:real_dataset_experiment}
For our real-dataset experiment, we draw on the dataset presented in \citep{yu2024aipatient}, which comprises approximately 1k conversations interactions with emergency hospital services. Each conversation is annotated to indicate its urgency level, following ESI. This dataset captures a variety of scenarios reflecting different degrees of criticality, thereby providing a realistic testbed for assessing how effectively the scheduling algorithm prioritizes urgent requests in a healthcare context.

\begin{figure}[!ht]
    \centering
    
    % Legend
    \includegraphics[height=.5in]{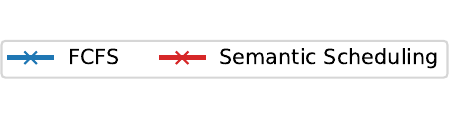}
    
    \vspace{1mm}
    
    % Two subfigures side by side
    \begin{subfigure}[t]{0.45\textwidth}
        \centering
        \includegraphics[height=2.0in]{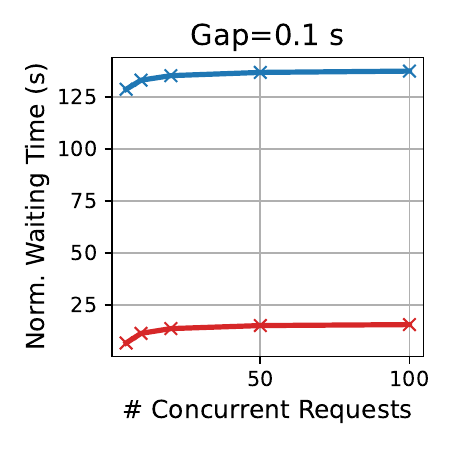}
        \caption{Gap=0.1s.}
        \label{fig:real_gap01}
    \end{subfigure}
    \hspace{5mm}
    \begin{subfigure}[t]{0.45\textwidth}
        \centering
        \includegraphics[height=2.0in]{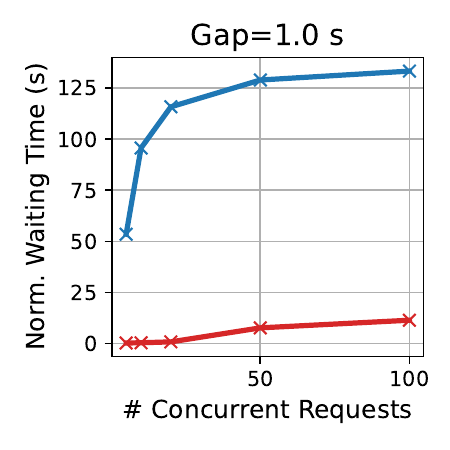}
        \caption{Gap=1.0s.}
        \label{fig:real_gap10}
    \end{subfigure}
    
    \caption{Scheduling performance on real-world dataset.}
    \label{fig:real_world_dataset_experiment}
\end{figure}

We examine requests of the highest urgency levels (Urgency 0) and investigate the normalized waiting time achieved by our semantic scheduling approach and the FCFS baseline on this real-world dataset (see~\Cref{fig:real_world_dataset_experiment}). We explore a wide range of configurations and test with the maximal number of concurrent requests ranging from 5 to 100 and different request gaps (0.1s and 1.0s). Our semantic scheduling approach  outperforms FCFS in all settings by achieving significantly lower normalized waiting time. With gap=0.1s, the normalized waiting time of our semantic scheduling approach is 6.69s when the maximal number of concurrent requests is 5, and slightly increases to 15.52s as the concurrency goes up to 100, which is up to 19.2x faster than FCFS.
With gap=1.0s, our semantic scheduling approach achieves as low as 0.32s normalized waiting time when concurrency is 5, a 167.3x speed-up compared to FCFS.

% \begin{figure}[!ht]
%     \centering
%     \begin{minipage}{0.45\textwidth}
%         \centering
%         \includegraphics[height=.33in]{Figures/Evaluation/legend_methods_real.pdf}
%     \end{minipage}
%     \vfill
%     \begin{subfigure}[t]{0.2\textwidth}
%     \includegraphics[height=1.2in]{Figures/Evaluation/real_gap0.1_all_a5000_4b.pdf}
%     % \caption{Gap=0.1s.}
%     % \label{fig:simulation_gap01}
%     \end{subfigure}%
%     ~
%     \begin{subfigure}[t]{0.2\textwidth}
%     \includegraphics[height=1.2in]{Figures/Evaluation/real_gap1.0_all_a5000_4b.pdf}
%     % \caption{Gap=1.0s.}
%     % \label{fig:simulation_gap099}
%     \end{subfigure}
%     \caption{Scheduling performance on real-world dataset.}
% \label{fig:real_world_dataset_experiment}
% \end{figure}

% Add to preamble:
% \usepackage{subcaption}

\section{Conclusions}
We presented a priority-aware serving system for large language models that addresses the critical challenge of providing differentiated service quality under resource constraints. By introducing a dual-heap architecture that jointly optimizes scheduling and memory management, our system ensures that high-priority requests receive preferential treatment in both computational ordering and cache retention. The asynchronous queue management design decouples request arrival from processing, enabling the system to handle burst traffic without sacrificing priority guarantees, while the stage-aware continuous batching mechanism prevents priority inversions between prefilling and decoding operations.

\appendix
\onecolumn
\section{More experiment results.}
\label{app:simulation_result}

\subsection{More Simulated Experiments}
\begin{figure}[!ht]
    \centering
    \begin{minipage}{\textwidth}
        \centering
        \includegraphics[height=.33in]{Figures/Evaluation/legend_methods.pdf}
    \end{minipage}
    \vspace{-2mm}
    \vfill
    % \begin{subfigure}[t]{0.2\textwidth}
    % \includegraphics[height=1.2in]{Figures/Appendix/simulation_gap0.1_all_a100_4b.pdf}
    % % \subcaption{Urgency.}
    % % \label{fig:simulation_priority}
    % \end{subfigure}%
    % ~
    \begin{subfigure}[t]{0.2\textwidth}
    \includegraphics[height=1.2in]{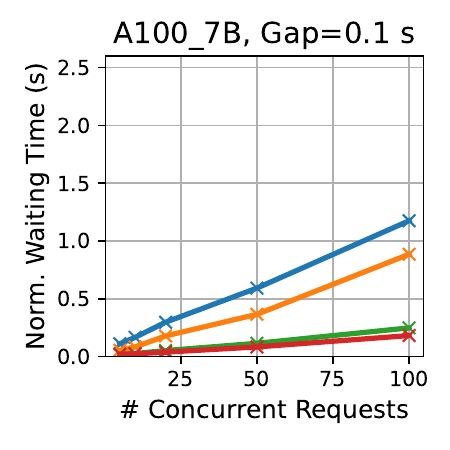}
    % \caption{Length.}
    % \label{fig:simulation_length}
    \end{subfigure}
    ~
    \begin{subfigure}[t]{0.2\textwidth}
    \includegraphics[height=1.2in]{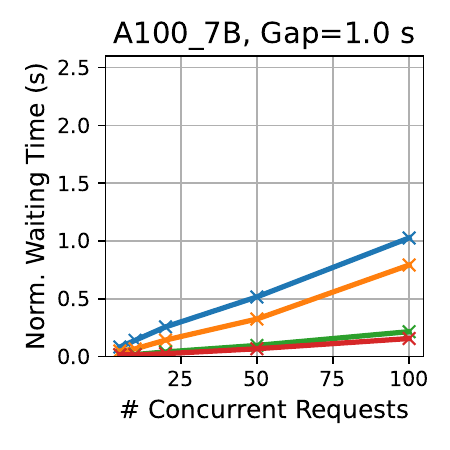}
    % \caption{Length.}
    % \label{fig:simulation_length}
    \end{subfigure}
    ~
    \begin{subfigure}[t]{0.2\textwidth}
    \includegraphics[height=1.2in]{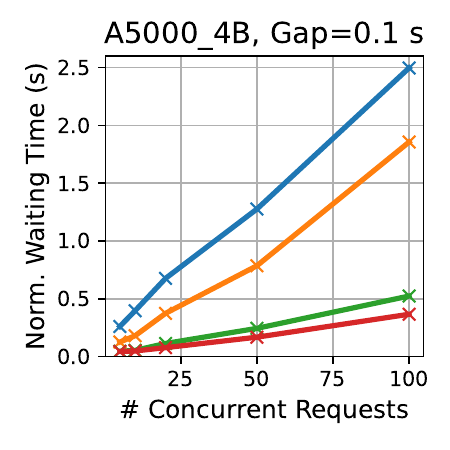}
    % \caption{Gap=0.1s.}
    % \label{fig:simulation_gap01}
    \end{subfigure}%
    % ~
    % \begin{subfigure}[t]{0.2\textwidth}
    % \includegraphics[height=1.2in]{Figures/Appendix/simulation_gap1.0_all_a100_4b.pdf}
    % % \subcaption{Urgency.}
    % % \label{fig:simulation_priority}
    % \end{subfigure}%
    ~
    \begin{subfigure}[t]{0.2\textwidth}
    \includegraphics[height=1.2in]{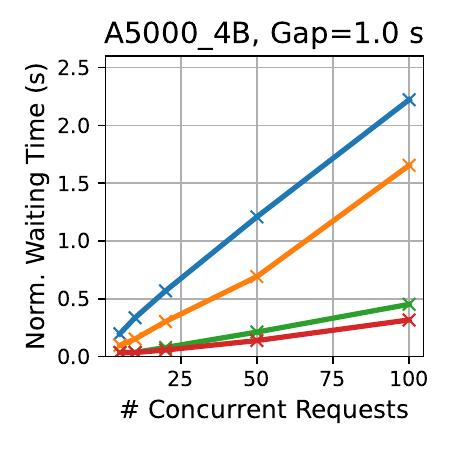}
    % \caption{Gap=0.1s.}
    % \label{fig:simulation_gap01}
    \end{subfigure}%
    \caption{More simulation results on the influence of spikes in user requests using Qwen1.5-7B on one NVIDIA A100 GPU and Qwen1.5-4B on one NVIDIA A5000 GPU.}
\label{fig:simulation_more_results}
\end{figure}

We further investigate the semantic scheduling performance under user request spikes in a wider range of LLM and GPU choices: Qwen1.5-7B on one NVIDIA A100 GPU and Qwen1.5-4B on one A5000 GPU.
We compare the achieved normalized waiting time for our semantic scheduling framework and all baselines -- FCFS, HPJF, and SJF.
We examine requests of the highest urgency levels (Urgency 0) and explore scenarios where requests may arrive at a high frequency (gaps vary from 0.1s to 1.0s and the maximal number of concurrent requests range from 5 to 100). We report results in~\Cref{fig:simulation_more_results} which resemble our analysis in~\Cref{subsec:simulated_experiments}. 
As the concurrent traffic increases, FCFS, SJF, and HPJF all struggle to process the large volume of incoming requests and incur a sharp increase in normalized waiting time for high-urgency requests, whereas semantic scheduling achieves the lowest normalized waiting time and outperforms all baselines.
% With the Qwen1.5-4B model on one NVIDIA A100 GPU, the normalized waiting time of semantic scheduling is up to 4.04s when gap=0.1 and 4.00s when gap=1.0s, which is up to 7.2x and 7.5x faster than the baselines.
Similarly, with the Qwen1.5-7B model on one NVIDIA A100 GPU, the normalized waiting time of semantic scheduling is up to 0.18s when gap=0.1 and 0.16s when gap=1.0s, which achieves speed-ups up to 6.5x and 6.6x compared to all the baselines.
Lastly, with the Qwen1.5-4B model on one NVIDIA A5000 GPU, the normalized waiting time of semantic scheduling is up to 0.37s when gap=0.1 and 0.32s when gap=1.0s, which achieves speed-ups up to 6.8x and 7.0x compared to all the baselines.

\subsection{More Real Dataset Experiments}
\begin{figure}[!ht]
    \centering
    \begin{minipage}{\textwidth}
        \centering
        \includegraphics[height=.33in]{Figures/Evaluation/legend_methods_real.pdf}
    \end{minipage}
    \vspace{-2mm}
    \vfill
    \begin{subfigure}[t]{0.2\textwidth}
    \includegraphics[height=1.2in]{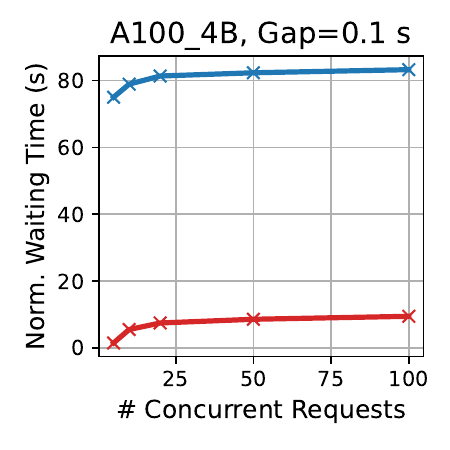}
    % \subcaption{Urgency.}
    % \label{fig:simulation_priority}
    \end{subfigure}%
    ~
    \begin{subfigure}[t]{0.2\textwidth}
    \includegraphics[height=1.2in]{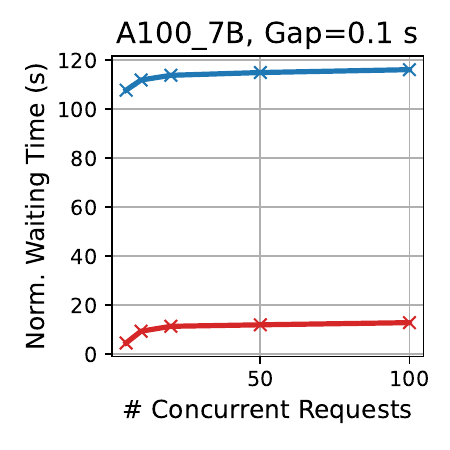}
    % \caption{Length.}
    % \label{fig:simulation_length}
    \end{subfigure}
    ~
    \begin{subfigure}[t]{0.2\textwidth}
    \includegraphics[height=1.2in]{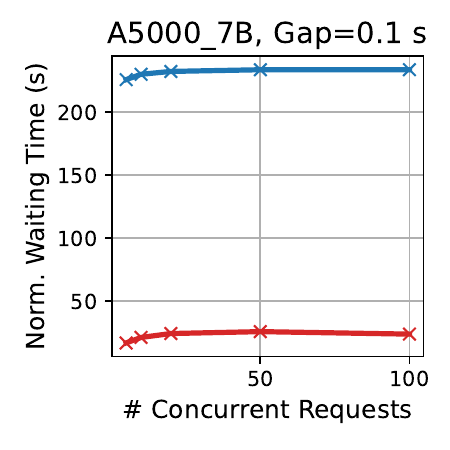}
    % \caption{Gap=0.1s.}
    % \label{fig:simulation_gap01}
    \end{subfigure}%
    \vfill
    \begin{subfigure}[t]{0.2\textwidth}
    \includegraphics[height=1.2in]{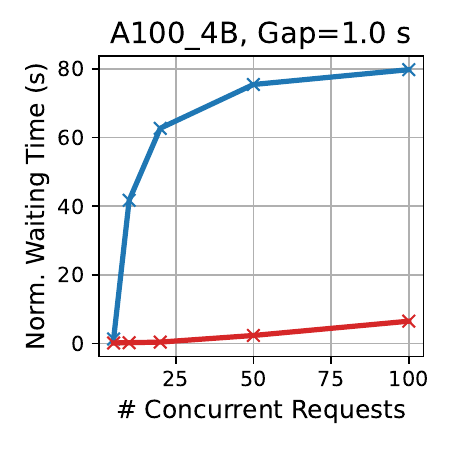}
    % \subcaption{Urgency.}
    % \label{fig:simulation_priority}
    \end{subfigure}%
    ~
    \begin{subfigure}[t]{0.2\textwidth}
    \includegraphics[height=1.2in]{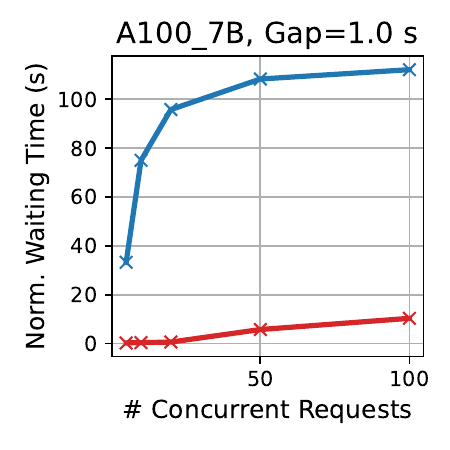}
    % \caption{Length.}
    % \label{fig:simulation_length}
    \end{subfigure}
    ~
    \begin{subfigure}[t]{0.2\textwidth}
    \includegraphics[height=1.2in]{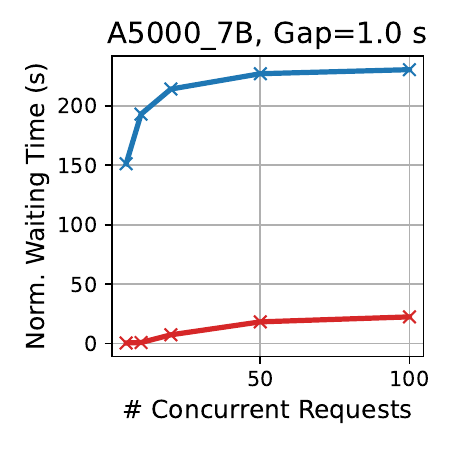}
    % \caption{Gap=0.1s.}
    % \label{fig:simulation_gap01}
    \end{subfigure}%
    \caption{More real-world evaluation results using Qwen1.5-4B and Qwen1.5-7B on one NVIDIA A100 GPU and A5000 GPU, separately.}
\label{fig:real_dataset_more_results}
\end{figure}

Similar to~\Cref{subsec:real_dataset_experiment}, we explore the capacity of the semantic scheduling performance on real-world datasets with a wide range of LLM and GPU choices:  Qwen1.5-4B and Qwen1.5-7B on one NVIDIA A100 GPU and A5000 GPU.
We compare the achieved normalized waiting time for our semantic scheduling framework and the FCFS baseline.
Results are summarized in~\Cref{fig:real_dataset_more_results}. 
As the concurrent traffic increases, semantic scheduling outperforms FCFS in all settings by achieving the lowest normalized waiting time.
With the Qwen1.5-4B model on one NVIDIA A100 GPU, the normalized waiting time of semantic scheduling is up to 9.47s when gap=0.1 and 6.51s when gap=1.0s, which is up to 49.7x and 166.7x faster than the FCFS baseline.
Similarly, with the Qwen1.5-7B model on one NVIDIA A100 GPU, the normalized waiting time of semantic scheduling is up to 12.87s when gap=0.1 and 10.32s when gap=1.0s, which achieves speed-ups up to 23.5x and 208.4x compared to the baseline.
Lastly, with the Qwen1.5-7B model on one NVIDIA A5000 GPU, the normalized waiting time of semantic scheduling is up to 23.80s when gap=0.1 and 22.47s when gap=1.0s, which achieves speed-ups up to 13.4x and 270.2x compared to the baseline.

\section{GPU Profile in Simulation Experiments}
\label{app:profile}
In this section, we specify the key coefficients and parameters used for our experimental settings based on different GPU configurations and model sizes. The settings include A100 with Qwen1.5-4B, A100 with Qwen1.5-7B, and A5000 with Qwen1.5-7B, where the coefficients define the prefill and decoding speeds, along with cache loading and saving speeds.

% For the A5000 with Qwen1.5-4B setting, the token decoding speed is determined by coefficients 
% 2.869$\times 10^{-6}$ and 3.008$\times 10^{-2}$, while the prefill speed uses coefficients 1.586$\times 10^{-9}$ and 1.305$\times 10^{-4}$. The cache loading and saving speeds are both 0.0003 per token.

For the A5000 with Qwen1.5-7B setting, the token decoding speed is determined by coefficients 2.117$\times10^{-6}$ and 2.727$\times10^{-2}$, and the prefill speed by 1.859$\times10^{-9}$ and 2.175$\times10^{-4}$. The cache loading and saving speeds are both 0.0003 per token.

For the A100 with Qwen1.5-4B setting, the token decoding speed coefficients are 5.913$\times10^{-9}$ and 1.196$\times10^{-2}$, and the prefill speed coefficients are 1.466$\times10^{-2}$ and 1.052$\times10^{-4}$. The cache loading and saving speeds are both 0.0001 per token.

For the A100 with Qwen1.5-7B setting, the token decoding speed coefficients are 1.349$\times10^{-8}$ and 1.330$\times10^{-2}$, while the prefill speed coefficients are 5.135$\times10^{-7}$ and 1.481$\times10^{-4}$. The cache loading and saving speeds are both 0.0001 per token.

These parameters were used to simulate realistic performance characteristics in the simulation experiments.

% You can have as much text here as you want. The main body must be at most $8$ pages long.
% For the final version, one more page can be added.
% If you want, you can use an appendix like this one.  

% The $\mathtt{\backslash onecolumn}$ command above can be kept in place if you prefer a one-column appendix, or can be removed if you prefer a two-column appendix.  Apart from this possible change, the style (font size, spacing, margins, page numbering, etc.) should be kept the same as the main body.
% %%%%%%%%%%%%%%%%%%%%%%%%%%%%%%%%%%%%%%%%%%%%%%%%%%%%%%%%%%%%%%%%%%%%%%%%%%%%%%%
% %%%%%%%%%%%%%%%%%%%%%%%%%%%%%%%%%%%%%%%%%%%%%%%%%%%%%%%%%%%%%%%%%%%%%%%%%%%%%%%

\section*{Impact Statement}
This paper introduces a semantic-aware scheduling framework for optimizing latency in LLM inference, prioritizing requests based on their urgency and computational requirements. Traditional scheduling approaches, such as FCFS, fail to account for the semantic significance of user queries, leading to inefficiencies in high-stakes domains such as emergency response, healthcare, financial services, and content moderation. Our method leverages semantic predictors and output length estimators to dynamically manage request prioritization, employing preemptive scheduling, KV cache optimization, and asynchronous queue management to reduce wait times for critical tasks.

Through both simulated and real-dataset experiments, we demonstrate significant reductions in response times for high-priority requests while maintaining overall system efficiency. By integrating machine learning-driven prioritization with incremental scheduling techniques, our approach establishes a foundation for more responsive and intelligent LLM serving systems, with broad applicability in time-sensitive AI deployment scenarios. This work paves the way for future advancements in context-aware scheduling algorithms, enabling LLMs to better serve critical real-world applications where urgency and fairness must be dynamically balanced.

\bibliography{references} 
\bibliographystyle{iclr2024_conference}

\end{document}